\documentclass[10pt,twocolumn,letterpaper]{article}

\usepackage[pagenumbers]{wacv} % To force page numbers, e.g. for an arXiv version

\usepackage{graphicx}
\usepackage{amsmath}
\usepackage{amssymb}

\usepackage{enumitem}
\usepackage{array}
\usepackage{pifont}
\usepackage[table]{xcolor}
\newcommand{\cmark}{\ding{51}}%
\newcommand{\xmark}{\ding{55}}%

\newcolumntype{C}[1]{>{\centering\arraybackslash}m{#1}}
\usepackage[export]{adjustbox}

\usepackage{cite}
\usepackage{array}
\usepackage{tabularx}
\usepackage{adjustbox}
\usepackage{tcolorbox}
\usepackage{amssymb}
\usepackage{float}

\usepackage{makecell}
\usepackage{amsmath}
\usepackage{color,soul}
\usepackage{mdframed}

\DeclareMathOperator{\cossim}{cossim}
\DeclareMathOperator{\MultiHead}{MultiHead}
\DeclareMathOperator{\mean}{mean}
\DeclareMathOperator{\Linear}{Linear}
\DeclareMathOperator{\ReLU}{ReLU}
\DeclareMathOperator{\Dropout}{Dropout}
\DeclareMathOperator{\source}{source}
\DeclareMathOperator{\VLM}{VLM}

\usepackage[pagebackref,breaklinks,colorlinks]{hyperref}

\usepackage[capitalize]{cleveref}
\crefname{section}{Sec.}{Secs.}
\Crefname{section}{Section}{Sections}
\Crefname{table}{Table}{Tables}
\crefname{table}{Tab.}{Tabs.}

\def\wacvPaperID{*****} % *** Enter the WACV Paper ID here
\def\confName{WACV}
\def\confYear{2025}

\begin{document}

\title{Zero-Shot Warning Generation for Misinformative Multimodal Content}

\author{Giovanni Delvecchio$^{1*}$, Huy H. Nguyen$^2$, and Isao Echizen$^{2,3}$ \\
\small{$^1$University of Bologna \ \ \ \ \ \ \ \ \ 
$^2$National Institute of Informatics (NII), Japan \ \ \ \ \ \ \ \ \ 
$^3$The University of Tokyo, Japan}\\
\small{*This work was carried out during the internship at NII, which was concluded on March 8, 2024.}\\
{\tt\small giovanni.delvecchio2@studio.unibo.it, \{nhhuy, iechizen\}@nii.ac.jp}}

\twocolumn[{
\maketitle
\begin{center}
    \centering
    \includegraphics[width=\textwidth]{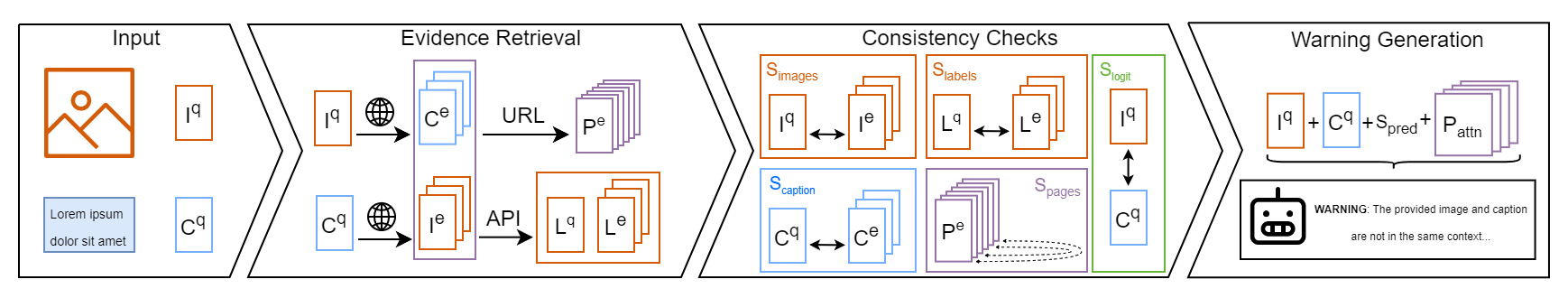}
    \captionof{figure}{Proposed pipeline for debunking misleading content. Each input image $I^{q}$ is used as a query to retrieve text $C^{e}$ from the web via inverse search, while each input caption $C^{q}$ retrieves images $I^{e}$ through direct search. Labels $L^{q}$ and $L^{e}$ are extracted using the Google Cloud Vision API. Consistency checks are needed to confront input data with evidence of the same type, in green we have the input image-caption consistency check. Each consistency check provides a consistency score; $S_{\text{pred}}$ is a vector containing all of them. The source pages $P^{e}$ of each piece of evidence are ranked to identify the most relevant pages $P_{\text{attn}}$. A frozen VLM with a custom prompt containing $I^{q}$, $C^{q}$, $S_{\text{pred}}$ and $P_{\text{attn}}$ generates an explanation contextualizing the input.}
    \label{fig:pipeline}
\end{center}
}]

%%%%%%%%% ABSTRACT
\begin{abstract}
   The widespread prevalence of misinformation poses significant societal concerns. Out-of-context misinformation, where authentic images are paired with false text, is particularly deceptive and easily misleads audiences. Most existing detection methods primarily evaluate image-text consistency but often lack sufficient explanations, which are essential for effectively debunking misinformation. We present a model that detects multimodal misinformation through cross-modality consistency checks, requiring minimal training time. Additionally, we propose a lightweight model that achieves competitive performance using only one-third of the parameters. We also introduce a dual-purpose zero-shot learning task for generating contextualized warnings, enabling automated debunking and enhancing user comprehension. Qualitative and human evaluations of the generated warnings highlight both the potential and limitations of our approach.
\end{abstract}

%%%%%%%%% BODY TEXT
% \blfootnote{The contribution of Giovanni Pio Delvecchio was made during the internship carried out at the National Institute of Informatics (Tokyo), which concluded on March 8, 2024.}

\section{Introduction} 
Misinformation has emerged as a topic of great concern in recent years, given its profound effect on both individuals and societies. At the individual level, the consequences of misinformation can manifest in local crime incidents involving conspiracy theorists~\cite{mcgonagle2017fake}. At the societal level, its effects permeate various domains including media (erosion of trust in news circulating on social media platforms), politics (damage to leaders' reputations), science (resistance to public health measures), and economics (influence on markets, consumer behavior and damage to brand reputation). 

The term ``misinformation" is often confused with related concepts such as fake news, disinformation, and deception. ``Misinformation" specifically refers to unintentional inaccuracies, such as errors in photo captions, dates, statistics, translations, or instances where satire is mistaken for truth. In contrast, disinformation involves the deliberate fabrication or manipulation of text, speech, or visual content as well as the intentional creation of conspiracy theories or rumors~\cite{hameleers2020picture}. Therefore, the key distinction between disinformation and misinformation lies in the intent behind sharing potentially harmful content. 

This study focuses on detecting out-of-context (OOC) image repurposing, a tactic used to support specific narratives~\cite{luo2021newsclippings}. Image repurposing is impactful as multimodal content combining text and images is more credible than text alone~\cite{hameleers2020picture} and easy to create. Our goal is to automate fact-checking by providing informative explanations to reconstruct the original context of an (image, caption) pair.

Figure~\ref{fig:pipeline} outlines the proposed pipeline in three steps: evidence retrieval, consistency check, and warning generation. Evidence retrieval involves searching web pages for the input image and comparing text spans, such as captions or titles, to the input caption. Similarly, captions help find other images for comparison. Source pages are analyzed for coherence and ranked by importance in deciding if the input pair is OOC. In warning generation, the Visual Language Model (VLM) MiniGPT-4\mbox{\cite{zhu2023minigpt}} provides either a contextual explanation or a warning about image repurposing, referencing relevant sources.

Our contributions are twofold: 
\begin{itemize}
    \item Proposing a flexible architecture for assessing input pair veracity by ranking evidence, achieving 87.04\% accuracy with the full model and 84.78\% with a lightweight version.
    \item Introducing a zero-shot learning task for warning generation, enabling debunking explanations with minimal computational resources.
    % \item Conducting ablations to evaluate the impact of redundant information, comparing backbones, and performing qualitative and human evaluations of generated warnings.
\end{itemize}

\section{Related Work}
\subsection{Closed Domain Approaches}
Liu et al.~\cite{liu2023robust} devised a system that leverages both domain generalization and domain adaptation techniques to mitigate discrepancies between hidden domains and reduce the modality gap, respectively.
Shalabi et al.~\cite{shalabi2024leveraging} addressed OOC detection by fine-tuning MiniGPT-4's alignment layer but without message generation and confidence scoring. Zhang et al.~\cite{zhang2023detecting} introduced a novel approach to reasoning over (image, caption) pairs. Instead of directly learning patterns from the data distributions, as Liu et al. did~\cite{liu2020visual}, they extract abstract meaning representation graphs from the captions and use them to generate queries for a VLM. This sophisticated approach enables a nuanced consistency check between the visual features of the image and the extracted features of the caption, but with limited room for explainability given by analysis of the generated queries and respective answers of the VLM. The work by Zhang et al.~\cite{zhang2023detecting} made it possible to understand the main limitation of closed-domain approaches to this task: evaluation of the veracity of an (image, caption) pair is sometimes challenging because the image may not depict all the statements that can be extracted from the caption. 

\subsection{Open Domain Approaches}
Popat et al.~\cite{popat2018declare} introduced the concept of detecting textual misinformation using external evidence; however,
in this study evidence is not integrated simultaneously.
Interpretability of the predictions is provided in the form of attention weights over the words of the analyzed document. Abdelnabi et al.~\cite{abdelnabi2022open} extended the concept of leveraging external knowledge for fact-checking to a multimodal (image, text) domain while also computing the aggregated consistencies considering all evidence at the same time for each of the two modalities.
A serious limitation of this approach is the provision of explanations solely in the form of attention scores signaling the most relevant evidence for the purpose of prediction along with limited debunking capabilities. 

Yao et al.~\cite{yao2023end} overcome this limitation by introducing an end-to-end pipeline consisting of evidence retrieval, claim verification, and explanation generation, using a dataset built from fact-checking websites with annotated evidence. A drawback of their approach is the utilization of a large language model (LLM) to summarize the evidence content, potentially overlooking important clues observable in the image.
Two parallel works, ESCNet~\cite{zhang2024escnet} by Zhang et al. and SNIFFER~\cite{qi2024sniffer} by Qi et al., also explore this area. ESCNet lacks explanation generation, while SNIFFER employs a commercial VLM (ChatGPT-4) for generating explanations.

Our work extends the principles outlined above~\cite{abdelnabi2022open, yao2023end} by introducing a module that reasons upon the source pages of the evidence and the generation of a contextualizing explanation as a zero-shot learning task of a VLM.

\section{Dataset and Evidence Collection} 

\subsection{NewsCLIPpings} In order to develop our contextualizing tool for the purpose of warning generation, 
we used the NewsCLIPpings dataset~\cite{luo2021newsclippings}, which is a synthetic dataset made by Luo et al.
built using the VisualNews~\cite{liu2020visual} corpus, which comprises news articles from
four prominent newspaper websites: The Guardian, BBC, USA Today, and The Washington Post.%\\
Given an (image, caption) pair, images of Visual News\mbox{\cite{luo2021newsclippings}} are retrieved and substituted to the original image to create falsified samples. Specifically, we used the merge-balanced subset, which consists of 71,072 training, 7,024 validation, and 7,264 test examples.
\subsection{External Evidence} The retrieved evidence were provided by Abdelnabi et al.~\cite{abdelnabi2022open}. %\\
%Given a pair $(I^{q}, C^{q})$, evidence obtained through means of direct search are the result of searching for images using $C^{q}$ as query (visual evidence); %\\
Given a pair $(I^{q}, C^{q})$, visual evidence $I^{e}$ is obtained through means of direct search using $C^{q}$ as query;
Textual evidence $C^{e}$ is obtained through means of inverse search, these are the result of searching for textual content (ad hoc scraped caption of images 
in web pages or title of those web pages) using $I^{q}$ as query.
Complementary informations, such as labels regarding the images $L^{q}$ and $L^{e}$, are obtained through means of Google Cloud Vision API\footnote{\url{https://cloud.google.com/vision/docs/labels}}. %\\
The authors of~\cite{abdelnabi2022open} provided the URLs to the source pages of both kind of evidence, which were downloaded for the purpose of this research.
\subsection{Sample Description} 
Each data point is described formally as
\begin{itemize}
\item query image $I^{q}$;
\item query caption $C^{q}$;
\item visual evidence retrieved using $C^{q}$ to query the web: $I^{e}=[I^{e}_{1}, ..., I^{e}_{n}]$;
\item labels obtained using Google Cloud Vision API: \\
$L^{q}$ are labels of $I^{q}$, and $L^{e}$ are labels of $I^{e}$;
\item textual evidence retrieved using $I^{q}$ to query the web (inverse search): \\
$C^{e}=[C^{e}_{1}, ..., C^{e}_{m}]$;
\item source pages $P^{e}=\source(I^{e})\cup \source(C^{e})$. 
\end{itemize}

\section{Proposed Method}
\begin{figure*}
\begin{center}
 \centering
 \includegraphics[width=0.80\textwidth]{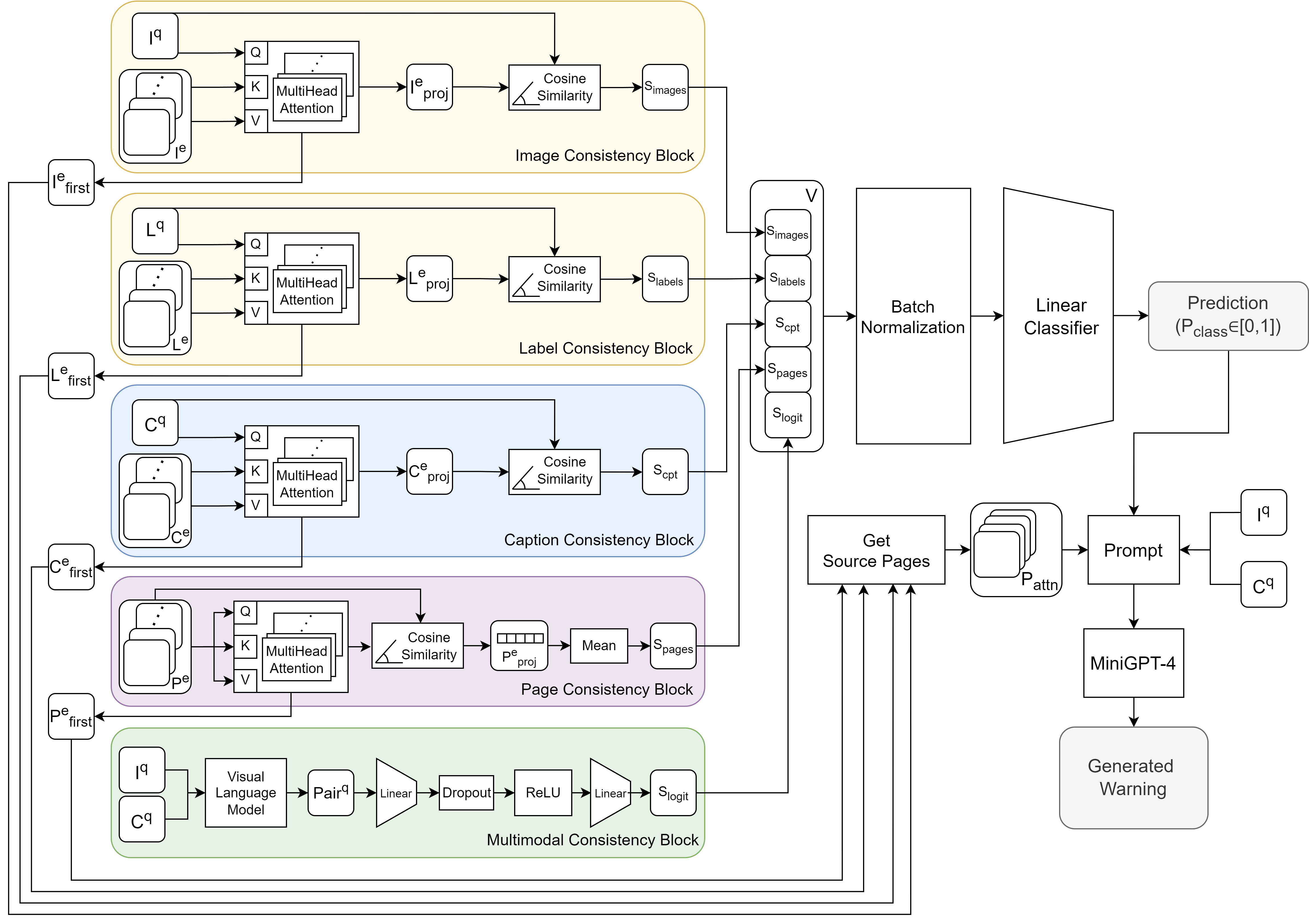}
 \vspace*{-4mm}
\caption{Overview of the proposed architecture for multimodal misinformation detection: Each consistency block provides a score in the range $[-1, 1]$ as it is the result of cosine similarity. The only exception is the score $S_{\text{logit}}$, which is the outcome of the multimodal consistency block (highlighted in green). The scores are concatenated, and the resulting vector is passed to the classification head, which consists of batch normalization~\cite{ioffe2015batch} and a linear layer. $P_{\text{attn}}$ are the source pages corresponding
to the evidence with the highest attention score from each attention-based block ($I^{e}_{\text{first}}$, $L^{e}_{\text{first}}$, $C^{e}_{\text{first}}$, $P^{e}_{\text{first}}$). The content of $P_{\text{attn}}$, together with the input pair and the final score $P_{\text{class}}$ are used to construct the input prompt of MiniGPT-4 for the purpose of warning generation.} 

 \label{fig:architecture}

 \end{center}
\end{figure*}

We designed an attention-based neural network capable of performing cross-consistency checks. In-depth analysis of the architecture, displayed in Figure~\ref{fig:architecture}, as well as the explanation generation task, are presented in the following subsections.

\subsection{Visual Reasoning} 
Images are represented either using ViT~\cite{wu2020visual} trained on ImageNet~\cite{deng2009imagenet} or DINOv2~\cite{oquab2023dinov2} embeddings for visual reasoning. These frozen visual transformers offer improved preservation of spatial information compared with ResNets~\cite{he2016deep, raghu2021vision}, which is crucial in our context. We aim to maintain structural information to achieve high similarity between images and their cropped/resized counterparts. Images are represented as vectors $I^{q},I^{e}\in\mathbb{R}^{768}$. The consistency score is computed as the cosine similarity:
\begin{center}
$\cossim(A, B) = \frac{\mathbf{A} \cdot \mathbf{B}}{\|\mathbf{A}\| \|\mathbf{B}\|}$\\
$S_{\text{images}}=\cossim(I^{q}, I^{e}_{\text{proj}})$\\
$\quad I^{e}_{proj} = \MultiHead(I^{q}, I^{e}, I^{e})$,
\end{center}
where $\MultiHead$ is the multi-head attention mechanism, as defined by Vaswani et al.~\cite{vaswani2017attention}.

\subsection{Textual Reasoning} 
We use frozen pre-trained sentence transformers, such as Sentence-BERT~\cite{reimers2019sentence} and Mini-LM~\cite{wang2020minilm}, for textual reasoning. The decision to use them was influenced by Nikolaev and Padó's research~\cite{nikolaev2023representation}, which showed that sentence transformers prioritize capturing semantic essence over grammatical functions or background details. This leads to increased cosine similarity between sentences with shared salient elements, like subjects or predicates. Labels and captions are represented as vectors $L^{q},L^{e},C^{q},C^{e}\in\mathbb{R}^{\text{ST-dim}}$, where $\text{ST-dim}$ is $768$ for Sentence-BERT and $384$ otherwise. The consistency scores are computed as
\begin{center}
$S_{\text{labels}}=\cossim(L^{q}, L^{e}_{\text{proj}})$\\ $\quad L^{e}_{\text{proj}} = \MultiHead(L^{q}, L^{e}, L^{e})$ \\
$S_{\text{cpt}}=\cossim(C^{q}, C^{e}_{\text{proj}})$ \\ $\quad C^{e}_{\text{proj}} = \MultiHead(C^{q}, C^{e}, C^{e})$.
\end{center}

\subsection{Page-Page Consistency Block} 
In this reasoning module, source pages are treated as text and represented as $P^{e}_{l}\in\mathbb{R}^{\text{ST-dim}}$. In particular, $P^{e}_{l}$ is computed as the mean of the embeddings of all sentences in the paragraphs on the $l-th$ source page, and $P^{e}_{l}$ is considered to be the embedding of the page itself. This module serves to identify the most important page for subsequent re-contextualization tasks and to compute an inter-agreement score between the retrieved documents by computing self-attention on page representations:
\begin{center}
$S_{\text{pages}} = \mean(\cossim(P^{e}, P^{e}_{\text{proj}}))$\\ $\quad P^{e}_{\text{proj}} = \MultiHead(P^{e}, P^{e}, P^{e})$.
\end{center}

\subsection{Multimodal Reasoning} 
We use late fusion~\cite{gadzicki2020early, pereira2023comparative} as the architectural pattern to address the lack of compelling explanations for the assessments made by existing detection methods. Our decision to use this pattern stems from both the flexibility it offers in swapping embeddings and the impracticality of training the entire model end-to-end due to its large parameter count. To address the challenges of inter-modality reasoning, we integrate a multimodal consistency block %(a.k.a. Image-Caption consistency block) 
that can be optimized alongside the rest of the model. To represent both $I^{q}$ and $C^{q}$ in the same latent space, we employ a VLM, chosen among two alternatives: CLIP~\cite{radford2021learning} with embedding dimension $\text{MM-dim}=512$ and MiniGPT-4~\cite{zhu2023minigpt} with $\text{MM-dim}=4096$. Each sample pair is represented as $Pair^{q}=\VLM(I^{q}, C^{q})\in\mathbb{R}^{\text{MM-dim}}$. The image-caption consistency block is defined as
\begin{center}
$S_{\text{logit}}=\Linear(\ReLU(\Dropout(\Linear(Pair^{q}))))$,
\end{center}
where the innermost linear layer projects $Pair^{q}$ into $\mathbb{R}^{256}$, the outermost linear layer projects it to $\mathbb{R}$, and the dropout probability is set to $0.2$.

\subsection{Classification Head} 
The output of the five consistency blocks is a single value, resulting in a vector $V\in\mathbb{R}^{5}$, which is fed into a \text{classification head} to obtain a prediction $P\in[0,1]$ (threshold $th$ set to 0.5 or found by analyzing the equal error rate, a pair is falsified for $P\geq th$, pristine otherwise). A batch normalization layer~\cite{ioffe2015batch} is added between $V$ and the \text{linear} classifier for faster convergence and higher accuracy. As the model is optimized for a binary classification task, binary cross-entropy loss is used:
\begin{center}
$l(x,y)=\mean(\{l_{1}, ..., l_{N}\}^{T}), $\\$\quad l_{n}=y_{n}\cdot \log x_{n}+(1-y_{n})\cdot \log (1-x_{n})$,
\end{center} 
where $x_{n}$ and $y_{n}$ are the predicted and ground truth labels, respectively, in a mini-batch of size $N=64$.

\subsection{Warning Generation} 
The last step of our pipeline consists of generating either contextual explanations for pristine pairs $(I^{q}, C^{q})$, providing additional insights about the depicted entities, or warnings indicating why $(I^{q}, C^{q})$ represent a case of image repurposing.
Given the versatility of LLMs in solving zero-shot learning tasks and the recent release of GPT-4~\cite{achiam2023gpt}, which extends these capabilities to a multimodal environment, we adopted a strategy for generating warnings without fine-tuning or reinforcement learning. 
Due to the unavailability of ground truth data for comparing explanations and optimizing the generation process, fine-tuning approaches are precluded by default.

Similarly, reinforcement learning through human feedback~\cite{ouyang2022training} presents challenges, including the need for a dedicated team to label examples and the potential persistence of model mistakes despite corrections. Reinforcement learning through AI feedback~\cite{lee2023rlaif}, which involves querying a larger model like GPT-4 to assess MiniGPT-4's outputs, also faces limitations. These include the constrained context window and the likelihood that all retrieved evidence may not contribute effectively to warning generation. Moreover, using a VLM to rectify errors made by another VLM inherently introduces comprehension discrepancies. Our approach to prompting was inspired by the work of Guo et al.~\cite{guo2023images}, who evaluated VLMs trained using multimodal pretraining similar to the Flamingo VLM~\cite{alayrac2022flamingo}, and is akin to the training strategy employed for MiniGPT-4.

MiniGPT-4 is the tool used for contextualization and is guided by the following prompt. Variables that depend on the values obtained by the consistency network are displayed between square brackets.

``\texttt{You are a tool for out-of-context detection, your task is to give reasons why the submitted image and the caption below
are in the same context or not.
Submitted Image: [$I^{q}$]. Caption: [$C^{q}$]. The likelihood of the submitted image and the above caption being in the same context is \textit{[$P_{\text{class}}$]}, thus the 
pair is \textit{[Falsified if $P_{\text{class}}\geq th$ else Pristine]}}".

Additionally, the title and the first few sentences (up to 400 characters)
of the 4 web pages with the highest attention score from each of the attention-based blocks  are included in prompt, together with the retrieval modality of the page, e.g.:

``\texttt{An evidence retrieved using the caption to query the web, obtained because it contains an image with high similarity with the submitted image has title \textit{[Title]} and content of the paragraphs \textit{[Content]}}". 

If the same source page is selected by multiple blocks, its title and content is added once to the prompt. For example, this redundancy could happen because a retrieved image and a retrieved caption belong to the same web page and both are ranked first by the multi-head attention mechanism (with reference to Fig. ~\ref{fig:architecture}, $\source(I^{e}_{\text{first}})=\source(C^{e}_{\text{first}})$).

\section{Experimental Analysis}
\subsection{Experimental Setup} All our experiments were carried out using either one or three NVIDIA A100 40-GB GPUs. Experiments on multiple GPUs were performed employing the distributed data parallel strategy. Early stopping was triggered if the validation loss stopped decreasing for 5 consecutive epochs. The designated mini-batch size was 64.
All our experiments employed a cyclic learning rate scheduler with initial learning rate equal to $9\cdot10^{-5}$ and maximum learning rate equal to $5\cdot10^{-4}$. Following the principle of scaling the learning rate with respect to the effective batch size, the learning rate was rescaled by multiplying both values by $1/\sqrt{3}$ when training on a single GPU.

\subsection{Comparison with State-of-the-Art Detectors}
NewsCLIPpings~\cite{luo2021newsclippings} established a baseline by fine-tuning CLIP (ViT/B-32)\cite{radford2021learning}, achieving 66.1\% accuracy. Shalabi et al.\cite{shalabi2024leveraging} reached state-of-the-art performance for closed-domain approaches by fine-tuning MiniGPT-4~\cite{zhu2023minigpt}, with 80.0\% accuracy.
Yao et al.\cite{yao2023end} (End-to-end in Table\ref{tab1}) employed text and image retrieval modules to select evidence, though these only consider the textual claim from the input, as their dataset relies on textual claims from fact-checking websites. Despite this limitation, their approach achieved 83.3\% accuracy on NewsCLIPpings.
Abdelnabi et al.~\cite{abdelnabi2022open} achieved 84.7\% accuracy using a 20.92M-parameter consistency-checking model trainable in 30 hours. Notably, CLIP is preliminarily fine-tuned (excluded from parameter and time counts).

Our lightweight model, employing frozen MiniGPT-4 as VLM, achieves 84.8\% of accuracy with 5.2 million parameters, it is trainable in 3 hours and 38 minutes on a single GPU, while it only requires 13 minutes on three GPUs. It is highlighted in blue in table~\ref{tab2}. It does not require any preliminary fine-tuning, there is no additional computation overhead due to comparison between entities and input caption, it leverages the analysis of source pages instead of exploiting domain representation and represents labels as text, comparing the labels of the query image with the labels of the visual evidence.
Our full-scale model achieves 87.04\% accuracy using the rescaled learning rate technique,
and 86.70\% using the standard learning rate. This model is highlighted in green in table~\ref{tab2}.
It counts 158 million parameters, of which 151 million belong to CLIP, which is optimized jointly with the rest of the architecture. The training time on a single GPU is 7 hours and 32 minutes, while on three GPUs it narrows to 27 minutes.

The parallel works, ESCNet~\cite{zhang2024escnet} and SNIFFER~\cite{qi2024sniffer}, demonstrate slightly higher performance than our approach (87.9\% and 88.4\%, respectively). However, ESCNet lacks explanation generation, while SNIFFER requires additional fine-tuning on the Q-Former in 16 hours. Furthermore, SNIFFER employs a different backbone (InstructBLIP), complicating a fair comparison.

\begin{table*}
%\begin{mdframed}[backgroundcolor=yellow!5, innertopmargin=2pt, innerbottommargin=2pt]
\caption{Comparison between our proposed approaches with others on the same dataset
(the merged/balanced split of the NewsCLIPpings\cite{luo2021newsclippings} dataset).}
\vspace*{-3mm}
\label{tab1}
% \hspace*{-0.8cm}
\begin{center}
\resizebox{0.7\textwidth}{!}{
\begin{tabular}
{|l|l|l|c|c|c|c|}
\hline
Model & Year & Paper & \begin{tabular}[c]{@{}c@{}}Requiring\\backbone\\fine-tuning\end{tabular} & \begin{tabular}[c]{@{}c@{}}Using\\reference\\sources\end{tabular} & \begin{tabular}[c]{@{}c@{}}Generating\\explanation\end{tabular} & \begin{tabular}[c]{@{}c@{}}Accuracy\\(\%)\end{tabular} \\
\hline
CLIP & 2021 & Luo et al.~\cite{luo2021newsclippings} & Yes & No & No & 66.1 \\
\rowcolor{gray!25} 
MiniGPT-4 & 2023 & Shalabi et al.~\cite{shalabi2024leveraging} & Q-Former & No & No & 80.0 \\
End-to-end & 2023 & Yao et al.~\cite{yao2023end} & Yes & \textbf{Yes} & Simple & 83.3 \\
\rowcolor{gray!25} 
CCN & 2022 & Abdelnabi et al.~\cite{abdelnabi2022open} & Yes & \textbf{Yes} & No & 84.7 \\
\textbf{Proposed (Lightweight)} & 2024 & Ours & \textbf{No} & \textbf{Yes} & \textbf{Yes} & 84.8 \\
\rowcolor{gray!25} 
\textbf{Proposed (Full-Scale)} & 2024 & Ours & \textbf{No} & \textbf{Yes} & \textbf{Yes} & 87.0 \\
ESCNet & 2024 & Zhang et al.~\cite{zhang2024escnet} & \textbf{No} & \textbf{Yes} & No & 87.9 \\
\rowcolor{gray!25} 
SNIFFER & 2024 & Qi et al.~\cite{qi2024sniffer} & Q-Former & \textbf{Yes} & \textbf{Yes} & 88.4 \\
\hline
\end{tabular}
}
\end{center}
%\end{mdframed}
\end{table*}

% table sep_______________________________
\begin{table*}
\centering
\caption{Classification performance, including ROC AUC and EER on validation set for different variants of the model; thEER is the threshold selected to minimize the EER on validation set. 
}\label{tab2}
% \hspace*{-2.4cm}
\resizebox{\textwidth}{!}{
\begin{tabular}{|C{1cm}|C{1cm}|C{1cm}|C{2cm}|C{1cm}|C{1cm}|C{1.5cm}|C{1.5cm}|C{1.5cm}|C{1.5cm}|C{1.5cm}|C{1.5cm}|} %|l|l|l|l|l|l|l|l|l|l|l|l|l|
\hline
%\# & 
\rotatebox{90}{Model Version   } &
\rotatebox{90}{Sentence Transformer  } & 
\rotatebox{90}{Vision Transformer  } & 
\rotatebox{90}{Multimodal Model  } & 
\rotatebox{90}{\makecell{Drop Labels-Labels\\Attention Block  }} &
\rotatebox{90}{\makecell{Drop Page-Page\\Self-Attention Block}} &
\rotatebox{90}{\makecell{Test Accuracy\\(th: 0.5)}} & 
\rotatebox{90}{thEER  } &  
\rotatebox{90}{\makecell{Test accuracy\\(thEER)}} & 
\rotatebox{90}{ROC AUC  } & 
\rotatebox{90}{EER  } & 
\rotatebox{90}{No. of Parameters  } \\
\hline
1 & std & std & CLIP & \xmark & \xmark & 86,46 & 0,5247 & 86,10 & 0,9312 & 0,1477 & 160 M \\
\rowcolor{gray!25} 
2 & std & std & CLIP & \cmark & \xmark & 86,43 & 0,5368 & 85,88 & 0,9329 & 0,1420 & 158 M \\
3 & std & std & CLIP & \xmark & \cmark & 85,79 & 0,5190 & 85,34 & 0,9285 & 0,1529 & 158 M \\
\rowcolor{gray!25} 
4 & std & std & CLIP & \cmark & \cmark & 86,31 & 0,5375 & 85,90 & 0,9249 & 0,1549 & 156 M \\
\hline

5 & alt & alt & CLIP & \xmark  & \xmark & 86,31 & 0,5515 & 86,05 & 0,9308 & 0,1463 & 155 M \\
\rowcolor{gray!25} 
6 & alt & alt & CLIP & \cmark  & \xmark & 86,41 & 0,5484 & 86,26 & 0,927  & 0,1492 & 154 M \\
7 & alt & alt & CLIP & \xmark  & \cmark & 86,21 & 0,5516 & 85,63 & 0,9273 & 0,1512 & 154 M \\
\rowcolor{gray!25} 
8 & alt & alt & CLIP & \cmark  & \cmark & 86,08 & 0,5577 & 85,76 & 0,9274 & 0,1477 & 154 M \\
\hline

9 & std & alt & CLIP & \xmark  & \xmark & 86,6  & 0,5451 & 86,26 & 0,932  & 0,1443 & 160 M \\
\rowcolor{green!25} 
10 & std & alt & CLIP & \cmark  & \xmark & 86,70 & 0,5522 & 86,5  & 0,9315 & 0,1391 & 158 M \\
11 & std & alt & CLIP & \xmark  & \cmark & 86,14 & 0,5400 & 86,31 & 0,9291 & 0,1452 & 158 M \\
\rowcolor{gray!25} 
12 & std & alt & CLIP & \cmark  & \cmark & 86,53 & 0,5583 & 85,59 & 0,9293 & 0,1452 & 156 M \\
\hline

13 & std & std & MiniGPT-4 & \xmark  & \xmark & 84,31 & 0,5168 & 84,25 & 0,9193 & 0,1621 & 10,5 M \\
\rowcolor{gray!25} 
14 & std & std & MiniGPT-4 & \cmark  & \xmark & 83,90 & 0,5171 & 84,10 & 0,9193 & 0,1612 & 8,1 M  \\
15 & std & std & MiniGPT-4 & \xmark  & \cmark & 84,03 & 0,4914 & 84,03 & 0,9193 & 0,1606 & 8,1 M  \\
\rowcolor{gray!25} 
16 & std & std & MiniGPT-4 & \cmark  & \cmark & 83,81 & 0,511  & 83,44 & 0,9179 & 0,1650 & 5,8 M \\
\hline
\rowcolor{cyan!25} 
17 & alt & alt & MiniGPT-4 & \xmark  & \xmark & 84,78 & 0,5446 & 84,60 & 0,9196 & 0,1615 & 5,2 M \\
\rowcolor{gray!25} 
18 & alt & alt & MiniGPT-4 & \cmark  & \xmark & 84,29 & 0,5464 & 84,25 & 0,9190 & 0,1578 & 4,6 M \\
19 & alt & alt & MiniGPT-4 & \xmark  & \cmark & 84,63 & 0,5526 & 84,01 & 0,9154 & 0,1658 & 4,6 M \\
\rowcolor{gray!25} 
20 & alt & alt & MiniGPT-4 & \cmark  & \cmark & 83,89 & 0,5521 & 83,49 & 0,9161 & 0,1627 & 4,0 M \\
\hline

21 & std & alt & MiniGPT-4 & \xmark  & \xmark & 84,65 & 0,5668 & 84,28 & 0,9209 & 0,1615 & 10,5 M \\
\rowcolor{gray!25} 
22 & std & alt & MiniGPT-4 & \cmark  & \xmark & 84,54 & 0,5513 & 84,29 & 0,9214 & 0,1589 & 8,1 M  \\
23 & std & alt & MiniGPT-4 & \xmark  & \cmark & 84,73 & 0,5677 & 83,96 & 0,9197 & 0,1609 & 8,1 M  \\
\rowcolor{gray!25} 
24 & std & alt & MiniGPT-4 & \cmark  & \cmark & 84,16 & 0,5743 & 83,85 & 0,9185 & 0,1652 & 5,8 M \\
\hline
\end{tabular}
}
\end{table*}

\subsection{Transformers Selection} 
The standard (\texttt{std}) and alternative (\texttt{alt}) sentence transformers are, respectively, Sentence-BERT\footnote{\url{https://huggingface.co/sentence-transformers/paraphrase-mpnet-base-v2}}\cite{reimers2019sentence} and a fine-tuned version of MiniLM\footnote{\url{https://huggingface.co/sentence-transformers/all-MiniLM-L6-v2}}\cite{wang2020minilm}. The key differences between them, in the context of our research, are that the former outputs embeddings that are twice as large as those output by the latter (768 against 384) and that the latter is faster at inference time and also more task specific because its training recipe is based on distillation and it is fine-tuned on sentences and small paragraphs.
The \texttt{std} and \texttt{alt} vision transformers are ViT\footnote{\url{https://huggingface.co/google/vit-base-patch16-224-in21k}}\cite{wu2020visual} trained on ImageNet-21k and DINOv2\footnote{\url{https://huggingface.co/docs/transformers/en/model\_doc/dinov2}}\cite{oquab2023dinov2}. These transformers share the same embedding dimension (768), and, analogous to the two sentence transformers, the latter has a different training recipe, which includes self-supervised learning on a larger corpus of uncurated data, enabling the extraction of visual features that work across image distributions and tasks. Our experiments demonstrated that the \texttt{std} sentence transformer and \texttt{alt} vision transformer was the best combination, regardless of the choice of multimodal model, when considering all four design patterns involving the removal of attention blocks. In fact, model version 17 had accuracy scores comparable to those of version 21 and a slightly lower ROC AUC, while versions 18, 19, and 20, compared with 22, 23, and 24, had lower accuracies (th:05) and lower ROC AUC. Nevertheless, we consider version 17 to be the best model version using the MiniGPT-4 model as it had the highest accuracy and fewer parameters while using all our attention blocks.

\subsection{Ablation Study: Block Suppression}

With the aim of understanding the true utility of labels, we treat them as sentences and dedicate a separate attention block to %them. 
evaluate their consistency.
With the CLIP backbone, model pair (1, 2) had comparable performance with and without the label-label attention block, whereas for pairs (5, 6) and (9, 10), dropping this block of redundant information resulted in slightly better performance. For the MiniGPT-4 versions, dropping this block meant removing a non-negligible number of parameters. The resulting decrease in performance observable in model pairs (13, 14), (17, 18), and (21, 22) supports our hypothesis that, in this regime, additional information about labels is helpful in achieving better performances.
We rehydrated the dataset by retrieving the source pages of each evidence.
The embeddings of source pages $P^{e}$ were then used in the self-attention layer with the aim of extracting the most relevant web page for the purpose of prediction, potentially enhancing explainability. Although dropping the page-page attention block tended to reduce performance, model version 23 performed optimally at th=0.5. Nonetheless, the other metrics suggest that versions 21 and 22, which use this block, performed better.

\subsection{Human Evaluation} 
Since the ground truth was unavailable for performance evaluation, unlike with the approach taken by Yao et al.~\cite{yao2023end}, we randomly selected 100 samples from the test set. These samples were evaluated by 20 individuals, each assessing 5 samples. Two evaluation metrics were used in accordance with the work of van der Lee et al.~\cite{van2021human}: `Informativeness' and `Overall Quality', both ranging from 0 to 5 with steps of 1. The `Informativeness' score is defined as the relevance and correctness of the output relative to the input specification. Evaluators considered both the generated warnings and the associated links when assigning the Informativeness score. The `Overall Quality' metric represents a judgment regarding the system's performance across the five samples observed by each evaluator. The average `Informativeness' score was \textbf{3.5 out of 5,} and the average Overall Quality was \textbf{4 out of 5}. These high scores can be attributed to the relatively straightforward contextualization of truthful examples when the correct evidence is retrieved, while various factors can render a warning unreliable in the case of falsified examples, leading to lower scores, especially if valuable evidence is not retrieved.

%table 4
\begin{table*}[h!]
%\vspace{-4.5cm}
\caption{Qualitative analysis of warning generation on one pristine sample (first row) and three falsify samples (remaining rows). The first and second one were contextualized correctly while the third and fourth one was contextualized incorrectly. In the fourth one, the selected evidence is not correlated to $I^{q}$ and $C^{q}$.  
}\label{tab3}
% \hspace*{-3.8cm}
\resizebox{\textwidth}{!}{
\begin{tabular}{|C{6cm}|m{10cm}|m{7cm}|}
%{|C{5cm}|p{8cm}|p{5cm}|} %|l|l|l|l|l|l|l|l|l|l|l|l|l|

\hline
Sample & 
\multicolumn{1}{|>{\centering\arraybackslash}m{10cm}|}{Generated Warning}  & 
\multicolumn{1}{|>{\centering\arraybackslash}m{7cm}|}{Retrieved Links} \\
\hline

\vspace{0.5cm}
%\begin{tcolorbox}[colback=red!5!white, colframe=red!75!black, left=-0.5mm, width=47mm]
\begin{tcolorbox}[colback=green!5!white, colframe=green!75!black, left=-0.5mm, width=57mm]
	\begin{minipage}{5.5cm}
		\begin{center}
			\includegraphics[width=50mm, margin=0pt 0ex 0pt 0mm]{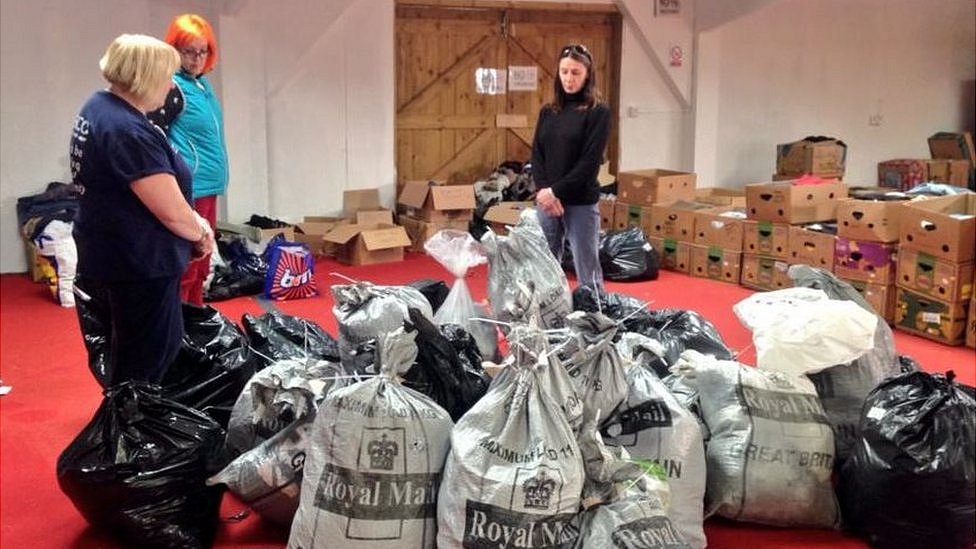}
		\end{center}
		Volunteers at Wadebridge Cornish Refugee Aid Collection in Cornwall stopped sorting through donations to observe the silence.
	\end{minipage} 
\end{tcolorbox}
&
\vspace{-2.2cm}\centering \begin{minipage}{9.5cm}
	CONTEXT: The image and the caption are in the same context as they both depict a group of people observing a minute's silence in memory of victims. 
	The image shows volunteers at Wadebridge Cornish Refugee Aid Collection in Cornwall stopping their work to observe the silence, while the caption describes a similar moment of silence observed across the UK in solidarity with the victims of the Paris terror attacks. 
	The evidence retrieved from the web provides further support for the pair being in the same context. 
	The article ``Paris attacks: Minute's silence observed in memory of victims - BBC News" describes a one-minute silence held across the UK to coincide with tributes in France and Europe following the attacks on bars, restaurants, a concert hall, and the Stade de France. 
	This is consistent with the image and caption depicting a moment of silence observed by volunteers at a refugee aid collection in Cornwall. 
	Overall, the likelihood of the image and caption being in the same context is 0.8520101308822632, indicating that they are pristine.
	\vspace{0.2 cm}
\end{minipage}
& 
%\begin{minipage}{5cm}
%\texttt{\url{https://www.bbc.com/news/uk-34831010?ns\_mchannel=social\&}}
%\end{minipage}  
\begin{center}
	\begin{minipage}{6.5cm}
		\vspace{0cm}
		\url{https://www.bbc.com/news/uk-34831010?ns_mchannel=social}
		%\href{https://www.bbc.com/news/uk-34831010?ns_mchannel=social}
		%{\texttt{https://www.bbc.com/news\\/uk-34831010?\\ns\_mchannel=social}}
	\end{minipage}
\end{center}
\\
\hline

\vspace{0.2cm}
%\begin{tcolorbox}[colback=red!5!white, colframe=red!75!black, left=-0.5mm, width=47mm]
\begin{tcolorbox}[colback=red!5!white, colframe=red!75!black, left=-0.5mm, width=57mm]
\begin{minipage}{5.5cm}
\begin{center}
\includegraphics[width=50mm, margin=0pt 0ex 0pt 0mm]{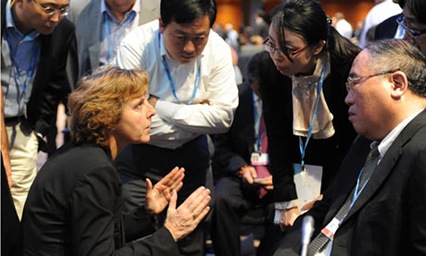}
\end{center}
Indian environment minister Jairam Ramesh left and China s senior climate change adviser Xie Zhenhua centre left with South African and Brazilian representatives in Delhi.
\end{minipage} 
\end{tcolorbox}
\vspace{0.2 cm}
&
\vspace{-5cm}\centering \begin{minipage}{9.5cm}
WARNING: The image and caption you provided are not in the same context. The image depicts a group of people in a conference room, while the caption describes a meeting of ministers from four emerging economies, including China, India, Brazil, and South Africa. The two elements do not match, and therefore, the pair is falsified. To support this conclusion, I have retrieved two web evidence using the caption and image respectively. The first evidence obtained through the caption shows that the meeting described in the caption took place in Delhi, while the image depicts a different scene. The second evidence obtained through the image shows that it is a generic image of a conference room, and not related to the meeting described in the caption. Therefore, based on these evidence, it can be concluded that the image and caption are not in the same context, and the pair is falsified.
\end{minipage}
& 

\vspace{-3cm}
\begin{adjustbox}{margin=0ex 0ex 0ex 0ex}
\begin{minipage}{6.5cm}
\begin{enumerate}[leftmargin=*]
    \item \url{https://www.theguardian.com/environment/2010/jan/24/china-climate-change-adviser}
    \item \url{https://www.theguardian.com/environment/cop18-doha-climate-change-conference}
\end{enumerate}
\end{minipage}
\end{adjustbox}
\\
\hline
% second row
\vspace{0.2 cm}
%\begin{tcolorbox}[colback=red!5!white, colframe=red!75!black, left=-0.5mm, width=47mm]
\begin{tcolorbox}[colback=red!5!white, colframe=red!75!black, left=-0.5mm, width=57mm]
\begin{minipage}{5.5cm}
\begin{center}
\includegraphics[width=50mm, margin=0pt 0ex 0pt 0mm]{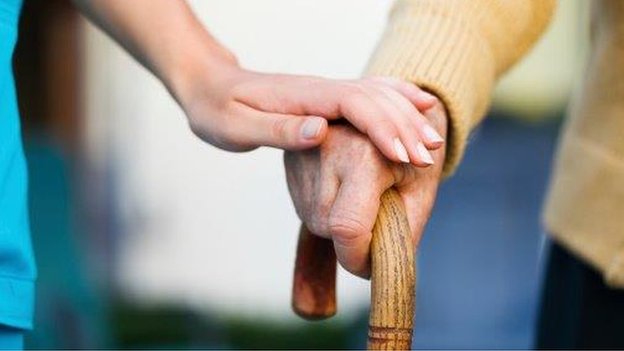}
\end{center}
At least 23 of the victims came from Britain Canada New Zealand Australia and Belgium.
\end{minipage}
\end{tcolorbox}
\vspace{0.2 cm}
&
\vspace{-3cm}\centering \begin{minipage}{9.5cm}
WARNING: The image and caption provided are not in the same context. The image depicts an elderly person holding a cane, while the caption provides information about the home front during World War II. There is no obvious connection between the two, and the likelihood of them being in the same context is low. The image and caption were retrieved from different sources, with the image coming from a stock photo website and the caption being a summary of information from Wikipedia. While the image may be related to elderly care, there is no evidence to suggest that it is specifically related to the home front during World War II. Therefore, I cannot provide any further information or insights related to this pair of image and caption.
\end{minipage}
& 
\vspace{-3cm}
%\hspace*{2cm}
\begin{adjustbox}{margin=0ex 0ex 0ex 0ex}
\begin{minipage}{6.5cm}
\begin{enumerate}[leftmargin=*]
    \item \url{https://en.wikipedia.org/wiki/Hawker_Hurricane}
    \item \url{https://www.bbc.com/news/world-asia-china-51403795}
    \item \url{https://en.wikipedia.org/wiki/Home_front_during_World_War_II}
    \item \url{https://www.entrepreneur.com/article/312224}
\end{enumerate}
\end{minipage}
\end{adjustbox}
\\
\hline

%third row
\vspace{0.1cm}
%\begin{tcolorbox}[colback=red!5!white, colframe=red!75!black, left=-0.5mm, width=47mm]
\begin{tcolorbox}[colback=green!15!white, colframe=red!75!black, left=-0.5mm, width=57mm]
\begin{minipage}{5.5cm}
\begin{center}
\includegraphics[width=50mm, margin=0pt 0ex 0pt 0mm]{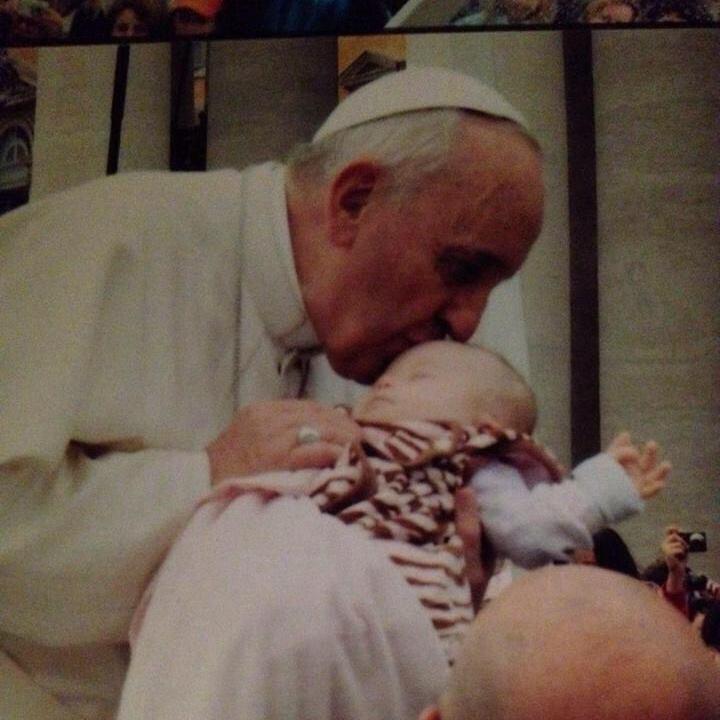}
\end{center}
Pope Francis kisses a young child with Down syndrome in Rome.
\end{minipage} 
\end{tcolorbox}
\vspace{0.2 cm}
&
\vspace{-3cm}\centering \begin{minipage}{9.5cm}
CONTEXT: The image depicts Pope Francis kissing a young child with Down syndrome in Rome. The caption describes the event as the child's wish to meet the Pope coming true. The evidence retrieved from the web, titled ``Young Ohio boy's wish to meet Pope Francis comes true - The Dialog" supports the context of the image and caption. The article describes a 12-year-old boy named Peter Lombardi who traveled to Rome to meet Pope Francis and was granted the opportunity to ride in the popemobile with him. The image and caption match the described event in the article, further indicating that they are in the same context. Therefore, the likelihood of the submitted image and the above caption being in the same context is 0.9954937100410461, which is very high, suggesting that the pair is pristine.
\end{minipage}
& 

\vspace{-1cm}
\begin{center}
\begin{minipage}{6.5cm}
\vspace{0.5cm}
\url{http://thedialog.org/featured/young-ohio-boys-wish-to-meet-pope-francis-comes-true/}
\end{minipage}
\end{center}

\\
\hline

\end{tabular}
}
\end{table*}
%\FloatBarrier
%\vspace*{-1cm}
%conclusion was here
\subsection{Qualitative Analysis} 
In Table~\ref{tab3}, we present samples used in the human evaluation. The ``Generated Warning" and ``Retrieved Links" columns display the output of the system. In the first two samples, coherent explanations and links are displayed. In the third sample, although the prediction was correct, the VLM failed to distinguish events even when different dates were provided by the caption and the evidence. Additionally, the system focused on why certain evidence had been retrieved instead of generating an explanation based on the content of the source pages. The third sample was also correctly predicted; however, $C^{q}$ is missing from the evidence. Consequently, the explanation provided is inconsistent with the source page of the caption, as evident from the presence of noisy evidence. Specifically, the warning explains that the reported victims in $C^{q}$ were victims of war, whereas the source page indicates that they were victims of child exploitation. Similarly, in the last sample, $I^{q}$ is missing from the retrieved evidence. Consequently, the pair is misclassified as pristine even though the child in $I^{q}$ was not affected by Down syndrome. However, the evidence contains a picture of the actual subject of $C^{q}$, making it simple for a human to distinguish this sample as OOC regardless of the generated explanation. 

\section{Discussion and Limitations}
Our proposed system relies heavily on search engine results, which may present conflicting evidence. However, addressing such conflicts would necessitate data manipulation beyond the scope of this work. While incorporating context from the source page in the reasoning process helps alleviate the issue of evidence resemblance across truthful and falsified examples, identifying the actual distinction between similar evidence remains a challenge. Incorporating labeled evidence or establishing a ground truth for generated warnings would enhance performance, both quantitatively and qualitatively. Additionally, evaluating the trustworthiness of the information sources would be beneficial in further enhancing the system's capabilities.

\section{Conclusion}
Our proposed system for detecting misinformative multimodal content leverages the masked multihead attention mechanism~\cite{vaswani2017attention} to increase the number of consistency-checking blocks while preserving conceptual simplicity. It demonstrated notable improvements in accuracy %and ROC AUC 
along with reduced training time compared with the state-of-the-art models. %of Abdelnabi et al.~\cite{abdelnabi2022open}. 
Our lightweight alternative with substantially fewer parameters achieved comparable results. Integration of warning generation as a zero-shot learning task into our pipeline showcased promising performance in human evaluations. Despite limitations regarding the quality of search results, our system represents a significant advancement in automated fact-checking. It empowers journalists and individuals navigating information on platforms like social media to effortlessly determine the truth behind content, irrespective of the original intent behind its dissemination.

\section*{Acknowledgements}
This work was partially supported by JSPS KAKENHI Grants JP21H04907 and JP24H00732, by JST CREST Grants JPMJCR18A6 and JPMJCR20D3 including AIP challenge program, by JST AIP Acceleration Grant JPMJCR24U3, and by JST K Program Grant JPMJKP24C2 Japan.

%%%%%%%%% REFERENCES
{\small
\bibliographystyle{ieee_fullname}
\bibliography{ref}
}

\end{document}